\documentclass[runningheads]{llncs}
\usepackage[T1]{fontenc}
\usepackage{graphicx}
\usepackage{booktabs}
\usepackage[misc]{ifsym}

\usepackage{mwe}

\usepackage{graphicx}

\usepackage[pagebackref,breaklinks,colorlinks]{hyperref}

\newcommand{\projectwebsite}{\url{https://sites.google.com/view/xiper}}
\usepackage{algorithm}
\usepackage{algorithmic}
\usepackage{amsmath}
\usepackage{amsfonts}

\usepackage{caption}
\usepackage[export]{adjustbox}
\usepackage{cleveref}
\crefname{figure}{Fig.}{Figs.}

\begin{document}

\title{Reinforcement Learning from Cross-domain Videos with Video Prediction Model}

\titlerunning{Reinforcement Learning from Cross-domain Videos}

\author{Zhao Yang\inst{1} \and
Xinrui Zu\inst{1} \and
Jacob E. Kooi\inst{1} \and
Thomas Delliaux\inst{2} \and
He Liu\inst{1} \and
Shujian Yu\inst{1}\thanks{Corresponding author} \and
Kevin Sebastian Luck\inst{1} \and
Vincent Fran\c{c}ois-Lavet\inst{1}}

\institute{VU Amsterdam, The Netherlands\\
\email{z.yang3@vu.nl, x.zu@vu.nl, jacobkooi92@gmail.com,}\\
\email{h.liu17@student.vu.nl, s.yu3@vu.nl, k.s.luck@vu.nl,}\\
\email{vincent.francoislavet@vu.nl}
\and
ISAE-SUPAERO, France\\
\email{thomas.delliaux@isae-supaero.fr}}
\authorrunning{Zhao Yang et al.}



\maketitle              

\begin{abstract}
Reinforcement learning from expert videos across visually distinct domains is challenging due to the absence of reward signals and the presence of domain gaps. We introduce XIPER (Cross-domain Video Prediction Reward), a reward model for learning from expert videos collected in a visually different domain, where the agent's appearance differs due to factors such as color, morphology, or the sim-to-real gap. More specifically, XIPER trains a cross-domain video prediction model that maps agent observations into the expert domain and uses the prediction likelihood as a reward signal. Experiments on the DMC Color Suite (8 tasks) and DMC Body Suite (3 tasks) show that XIPER consistently outperforms baselines despite domain gaps such as differences in agent color and morphology. We further analyze XIPER on a sim-to-real transfer dataset, demonstrating that it produces meaningful reward signals for real-robot observations given only simulated expert videos. Code, pretrained models, datasets and video demonstrations can be found on our project webpage:~\projectwebsite.

\end{abstract}

\section{Introduction}
\label{sec:intro}

Humans and animals can learn useful behaviors by observing demonstrations~\cite{rizzolatti2004mirror,tenenbaum2011grow}, even when the demonstrator differs from the learner in visual appearance. For instance, pilots improve real-world performance by studying animated training videos~\cite{chenot2025assessing}. This is possible because the learner can extract the underlying behaviors from a demonstration without requiring an exact match in appearance. Endowing reinforcement learning (RL) agents with this same ability would unlock the vast pool of online videos as a source of expert demonstrations~\cite{schmeckpeper2020reinforcement,baker2022video,vuong2023open}, removing the need for laborious reward engineering~\cite{riedmiller2018learning} and costly domain-specific data collection~\cite{pertsch2021guided,rafailov2024d5rl}.

\begin{figure}[!htb]
    \centering
    \includegraphics[width=0.6\linewidth]{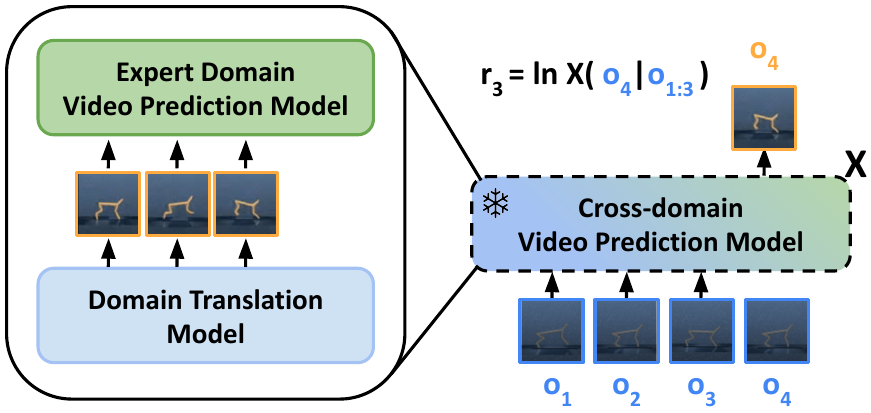}
    \caption{XIPER trains a cross-domain video prediction model to predict the next expert observation given a sequence of agent observations. The reward is calculated as the prediction likelihood, assigning higher rewards for agent observations consistent with expert videos.}
    \label{fig:concept}
\end{figure}

Achieving this in practice, however, remains challenging. Prior work~\cite{stadie2017third,cetin2021domain,choi2024domain} combines domain-invariant representation learning~\cite{zhao2019learning,li2021domain,nguyen2021domain} with adversarial training~\cite{goodfellow2014generative} to encourage behavioral similarity between the learner and the expert. However, such representations often fail to scale to complex scenarios, and adversarial methods remain susceptible to mode collapse~\cite{torabi2018generative,peng2021amp}.

Instead, our approach builds on recent progress in unpaired image-to-image translation~\cite{korotin2022neural} and generative modeling~\cite{yan2021videogpt,esser2021taming}. We introduce XIPER (Cross-domain Video Prediction Reward), a reward model that uses cross-domain video prediction likelihood as a training signal, bypassing the need for domain-invariant representations and adversarial training altogether.

XIPER trains a cross-domain video prediction model (as illustrated in~\cref{fig:concept}) comprising two components: (1)~a domain translation model that maps agent observations into the expert domain, and (2)~a video prediction model that predicts the next expert frame from a sequence of past ones. Both components are pretrained offline and frozen during RL training. At each time step, recent agent observations are first translated into the expert domain, then fed into the prediction model. The prediction likelihood serves as the reward signal: higher likelihoods indicate closer alignment with expert behavior, guiding the agent to reproduce demonstrated behaviors.

Concretely, our contributions are threefold:
\begin{itemize}
    \item We introduce XIPER, a cross-domain reward model that enables RL agents to learn from unlabeled expert videos across different domains, relying solely on pixel observations without ground-truth rewards or low-dimensional state information.
    \item We show that XIPER consistently outperforms three competitive baselines across the DMC~\cite{tunyasuvunakool2020dm_control} Color Suite (8 tasks) and DMC Body Suite (3 tasks), and analyze its applicability on a sim-to-real transfer dataset, in which expert videos are only available in simulation.
    \item We conduct ablation studies to analyze key design choices. Code, pretrained models, datasets and video demonstrations are provided on our project webpage.
\end{itemize}

\section{Related Work}
\label{sec:related_work}

XIPER addresses a setting where an RL agent learns from expert visual observations in a different domain, without access to action or reward labels. We review three lines of related work and position XIPER within each.

\subsection{Cross-domain RL from Visual Observations}
Reinforcement learning from cross-domain visual observations is challenging due to domain gaps~\cite{giammarino2024visually,raychaudhuri2021cross,choi2024domain,seo2022reinforcement} and the absence of action~\cite{foster2024behavior} and reward labels~\cite{park2025flow}. Prior work typically trains a domain-invariant discriminator to classify behaviors across domains and uses it as a reward model for adversarial imitation learning~\cite{torabi2018generative}. TPIL~\cite{stadie2017third} adds a domain confusion loss to encourage invariance, DisentanGAIL~\cite{cetin2021domain} improves upon this with mutual information constraints, and D3IL~\cite{choi2024domain} further applies cycle consistency objectives~\cite{zhu2017unpaired} at both image and feature levels. Despite these advances, discriminator-based approaches remain prone to mode collapse~\cite{jaegle2021imitation,zolna2021task} and unstable adversarial training~\cite{mescheder2018training}, and their policies still rely on ground-truth low-dimensional states rather than raw visual inputs. XIPER avoids adversarial training entirely by using video prediction likelihood as a reward signal, and operates directly on high-dimensional pixel observations, making the setting more realistic.

\subsection{Video Generation for Decision Making}
Recent work has explored using video generation models to either plan over future trajectories or provide reward signals for RL agents. UniPi~\cite{du2023learning} generates expert video trajectories with a text-conditioned diffusion model and infers actions via an inverse dynamics model. DR~\cite{Huang2023DiffusionReward} and VIPER~\cite{escontrela2024video} instead use video generation models to provide dense reward signals that encourage expert-like behavior. However, these methods assume the agent and expert share the same domain. XIPER extends this idea to the cross-domain setting by incorporating a domain translation step before video prediction, enabling reward generation even when the agent and expert domains differ.

\subsection{Image Translation in RL}
Image-to-image translation~\cite{kim2023unpaired,xie2023unpaired} has been widely applied in RL, primarily for policy transfer~\cite{kirk2021survey}. CycleGAN~\cite{zhu2017unpaired} has been used in Atari~\cite{gamrian2019transfer} to map distracted observations to a clean source domain, and RL-CycleGAN~\cite{rao2020rl} jointly optimizes translation and RL for sim-to-real transfer. RetinaGAN~\cite{ho2021retinagan} and \cite{yuan2022sim} similarly translate simulated images to the real world for policy deployment. AVID~\cite{smith2019avid} applies translation to human demonstration frames to generate robot reward signals. A common thread across these works is the assumption that expert policies are available, with translation serving mainly as a bridge for policy reuse. XIPER instead operates in a more challenging setting where only expert videos are available, integrating translation directly into a video prediction reward model to bridge domains without requiring access to policies or actions.

\section{Preliminaries and Problem Setup}
\label{sec:preliminaries}
\subsection{Preliminaries} 
We consider a Partially Observable Markov Decision Process (POMDP)~\cite{sutton1998introduction,franccois2018introduction}, defined by the tuple $\langle S, A, P, R, \Omega, O, \gamma \rangle$, where an agent interacts with an environment over discrete timesteps. At each timestep $t$, the environment is in a state $s_t \in S$, but the agent only receives a visual observation $o_t \in \Omega$ drawn from the observation function $O(o_t | s_t)$. The agent selects an action $a_t \in A$ according to the policy $\pi(a_t | o_t)$, receives a reward $r_t = R(s_t, a_t)$, and the environment transitions to the next state $s_{t+1} \sim P(\cdot | s_t, a_t)$. Here, $A$ is the action space, $S$ is the state space, $\Omega$ is the observation space, $O$ is the observation function, $P$ defines the transition dynamics, and $R$ is the reward function. In our setting, $R$ is unknown to the agent and is instead approximated by a learned reward model derived from expert demonstrations, as described in \cref{sec:problem_setup}. The agent's objective is to learn a policy $\pi$ that maximizes the expected cumulative discounted return:
\begin{equation}
    G_t = \mathbb{E}_{\pi} \left[ \sum_{k=0}^{T} \gamma^k r_{t+k} \right]
\end{equation}
where $\gamma \in [0,1]$ is the discount factor and $T$ is the task horizon.

\subsection{Problem Setup}
\label{sec:problem_setup}
We study the problem of visual imitation learning from unlabeled videos in a cross-domain setting, where the agent operates in a domain $\mathcal{A}$ using only visual observations $o_t$, without access to the underlying ground truth states $s_t$ or the environment's true reward function $R$. Instead, the agent is given expert visual observations $o_t'$ from a different domain $\mathcal{E}$. The demonstration videos differ from $\mathcal{A}$ in visual characteristics such as color schemes or agent morphologies, which are referred to as domain gaps~\cite{farhadi2024domain}, and do not have action or reward labels. 

Due to the absence of true reward signals in domain $\mathcal{A}$, an auxiliary reward function $R'$ must be derived from the unlabeled expert videos collected in domain $\mathcal{E}$. The core challenge lies in learning a reward function from these unlabeled videos that allows the agent to acquire expert-like behavior in a visually different environment. Since it is handy to perform random actions and collect these observations, we also assume having access to pre-collected random observations from both domains. More formally, given only:
\begin{itemize}
    \item Pre-collected expert and random observations (videos) from domain $\mathcal{E}$;
    \item Pre-collected random observations (videos) from domain $\mathcal{A}$,
\end{itemize}
we aim to learn a policy that enables the agent to imitate expert behavior in domain $\mathcal{A}$ from visual observations, despite the visual and semantic differences between domains. The key challenge is: how can we transfer knowledge from expert videos collected in one domain to train an agent that performs well in another, using only raw visual observation inputs and no access to rewards or states?

\section{Cross-domain Video Prediction Rewards}
\label{sec:method}

XIPER enables RL across visually distinct domains by learning a cross-domain video prediction model. Given a sequence of agent observations, the model predicts the next expert observation, and the likelihood of this prediction serves as a reward signal that encourages the agent to produce expert-like behaviors.

\subsection{Cross-domain Video Prediction Model}
The cross-domain video prediction model consists of two components (see~\cref{fig:concept}): (i) an expert video prediction model that captures the distribution of expert behaviors, and (ii) a domain translation model that maps agent observations into the expert domain. We describe each in turn.

\subsubsection{Expert Domain Video Prediction Model}
We train a VideoGPT~\cite{esser2021taming,yan2021videogpt} model on expert trajectories by optimizing the maximum likelihood objective:
\begin{equation}
    \max_{\theta} \sum_{t=1}^T \log p_\theta(o_t'|o_{1:t-1}'),
\end{equation}
where $p_\theta$ denotes the generative model parameterized by $\theta$. Since computing the likelihood over an entire video is expensive, we approximate it using a limited context window of $k=16$ frames, giving:
\begin{equation}
    \max_{\theta}\sum_{t=1}^T \log p_\theta(o_t'|o_{\max(1, t-k):t-1}').
    \label{eq:vp}
\end{equation}
This model assigns higher likelihoods to observation sequences that resemble expert behavior. Once trained, it provides an imitation reward based on how likely the agent's recent observations are under the expert distribution:
\begin{equation}
    r_t^{\text{imit}}=\log p_\theta (o_{t}'|o_{\max(1, t-k):t-1}').
\end{equation}
A higher $r_t^{\text{imit}}$ indicates that the recent trajectory more closely matches the expert distribution~\cite{escontrela2024video}. While we use VideoGPT here, other generative video models~\cite{yu2023magvit,melnik2024video} are also compatible.

\subsubsection{Domain Translation Model}
In a cross-domain setting, the expert video prediction model cannot directly process agent observations due to visual domain gaps. We address this with a domain translation model $T_\phi: \mathcal{A} \rightarrow \mathcal{E}$ that maps agent observations $o_t \in \mathcal{A}$ into the expert domain $o_t' \in \mathcal{E}$.

Since paired data across domains is unavailable, we formulate this as an unpaired image-to-image translation problem. Concretely, $T_\phi$ is trained using the Neural Optimal Transport (NOT)~\cite{korotin2022neural} objective, a neural approximation of the classical optimal transport problem, formulated as a min-max optimization:
\begin{equation}
\min_{\phi} \max_{\psi} \mathbb{E}_{o \sim \mu} \left[ f_\psi(T_\phi(o)) \right] - \mathbb{E}_{o' \sim \nu} \left[ f_\psi(o') \right] + \lambda \cdot \mathbb{E}_{o \sim \mu} \left[ c(o, T_\phi(o)) \right] \label{eq:ts}
\end{equation}
where $\mu$ and $\nu$ are distributions over agent and expert domain observations, $f_\psi$ is a critic function, and $c$ is a cost function. We refer the reader to~\cite{korotin2022neural} for further details. Other unpaired translation methods~\cite{zhu2017unpaired,kim2023unpaired} are equally compatible with our framework.

\subsubsection{Integrated Cross-Domain Model}
Composing the two components yields the cross-domain video prediction model $X_{\theta,\phi}$:
\begin{equation}
    o_{t}' = X_{\theta,\phi}(o_{t-k:t-1}) = p_{\theta}(T_\phi(o_{t-k:t-1})),
\end{equation}
which predicts the current expert observation given the previous $k$ agent observations by first translating them into the expert domain, then applying the video prediction model.

\subsection{Cross-domain Reward Formulation}
During training, the agent receives its current and previous $k$ observations from its own domain. These are first translated into the expert domain using $T_\phi$, and the XIPER reward is computed as the log-likelihood of the translated current observation given the translated context:
\begin{equation}
    r_t^{\text{xiper}}=\ln p_\theta (T_\phi(o_{t})|T_\phi(o_{t-k:t-1})).
\end{equation}
To encourage exploration, we augment this with an exploration bonus $r_t^{\text{expl}}$ following prior work~\cite{sekar2020planning,burda2018exploration,escontrela2024video}:
\begin{equation}
    \label{eq:reward}
    r_t=r_t^{\text{xiper}} + \beta \cdot r_t^{\text{expl}},
\end{equation}
where $\beta=1$ across all experiments. An ablation on $\beta$ is provided in~\cref{sec:ablation}.

\subsection{Data Curation}
XIPER requires three types of data: expert trajectories in domain $\mathcal{E}$ for training the video prediction model, and random observations from both domains for training the translation model. Expert trajectories are collected using pretrained RL agents, while random observations are gathered using uniformly sampled actions. As a result, XIPER assumes access to expert and random observations in domain $\mathcal{E}$, and only random observations in domain $\mathcal{A}$. In all experiments, 50 episodes are collected per task per domain.

\subsection{Implementation Details}
A high-level overview is provided in~\cref{alg:viper}. XIPER is built on top of VIPER~\cite{escontrela2024video}, which is video prediction reward model that is restricted to same-domain settings. XIPER generalizes this to cross-domain scenarios by incorporating domain translation into the reward pipeline, enabling reward generation from expert videos in a different domain. Concretely, XIPER uses DreamerV3~\cite{hafner2023mastering} as the RL backbone, a VideoGPT~\cite{yan2021videogpt} model trained on expert videos to capture expert behavior, and a NOT translation model~\cite{korotin2022neural} for domain bridging. Both the video prediction and translation models are frozen during policy learning. For the exploration bonus, we adopt the uncertainty-based objective from Plan2Explore~\cite{sekar2020planning}, which integrates naturally with DreamerV3. Code, pretrained models, datasets and video demonstrations are provided on our project webpage\footnote{~\projectwebsite}.

\begin{algorithm}
\caption{RL with XIPER Reward Signals}
\label{alg:viper}
\begin{algorithmic}[1]
\STATE Pretrain cross-domain video prediction model \( X_{\theta,\phi} \) on pre-collected expert and random videos with~\cref{eq:vp} and~\cref{eq:ts}. 
\STATE Initialize RL policy $\pi$, set up task environment env.
\WHILE{not converged}
    \STATE Choose action: $a_t \sim \pi(o_t)$
    \STATE Step environment:
     $o_{t+1} \leftarrow \text{env}(a_t)$
    \STATE Calculate reward: $r_t \leftarrow \ln X_{\theta,\phi}(o_{t+1} | o_{t-k:t}) + \beta r_t^{\text{expl}}$
    \STATE Add transition \((o_t, a_t, r_t, o_{t+1})\) to replay buffer
    \STATE Train $\pi$ from replay buffer with DreamerV3
\ENDWHILE
\STATE Output $\pi$ for evaluation.
\end{algorithmic}
\end{algorithm}

\section{Experiments}
\label{sec:experiments}

\begin{figure}[!htb]
    \centering
    \includegraphics[width=0.8\linewidth]{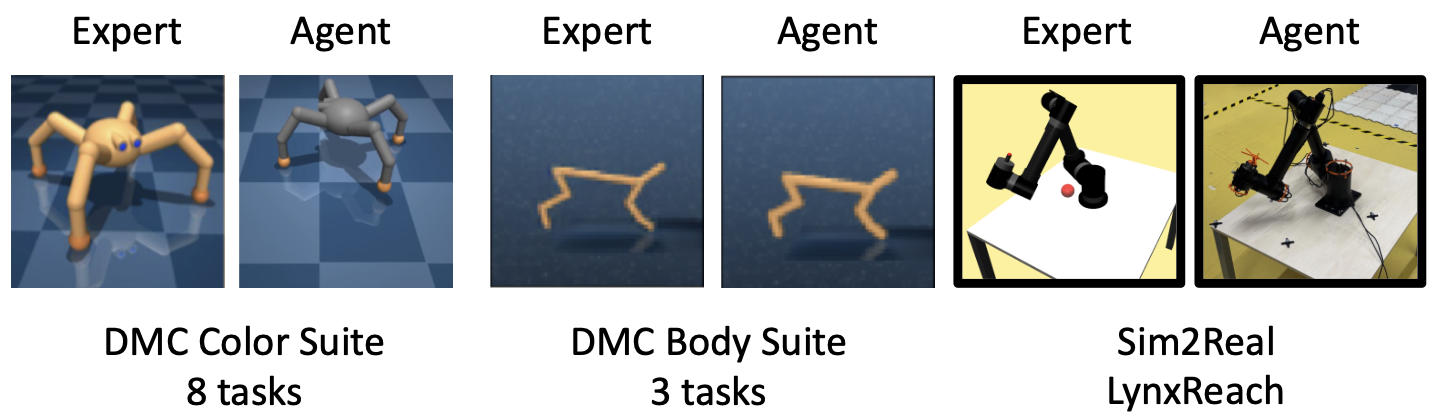}
    \caption{Tasks used in the experiments. The DMC Color Suite includes 8 tasks with the agent's color changed from orange to gray. The DMC Body Suite consists of 3 tasks where the agent's body shape is modified to be thicker. The Sim2Real analysis uses a Lynx Robot arm dataset, where expert videos are only available in simulation.}
    \label{fig:envs}
\end{figure}

We evaluate XIPER across two stages. First, we conduct a thorough simulation study addressing three questions: (1) How accurately does the cross-domain video prediction model capture expert behavior, and how well do its rewards correlate with ground-truth task rewards? (2) Are these learned rewards sufficient to train agents that acquire meaningful behaviors? (3) What are the key design choices and their impact on performance? We then conduct a dataset analysis in a sim-to-real setting, examining whether XIPER can generate meaningful reward signals for real-robot observations given only simulated expert videos (we defer the details to~\cref{sec:sim2real}).

For simulation experiments, we create two suites based on DMC~\cite{tunyasuvunakool2020dm_control} to evaluate all agents (see~\cref{fig:envs}):
\begin{itemize}
    \item DMC Color Suite: 8 tasks where expert data uses orange agents and the learning agent is gray, introducing a visual domain gap. Tasks: Quadruped-walk, Pointmass-easy, Cartpole-balance, Quadruped-run, Cup-catch, Pendulum-swingup, Cheetah-run, and Cartpole-swingup.
    \item DMC Body Suite: 3 tasks where the learning agent has a thicker body than the standard-shaped expert, introducing both visual and dynamics domain gaps. Tasks: Cartpole-balance, Cheetah-run, and Cartpole-swingup.
\end{itemize}

Most prior cross-domain visual RL methods rely on adversarial learning but use weak RL backbones~\cite{stadie2017third,cetin2021domain} that struggle with visually rich tasks. For fair comparison, we reimplement all baselines using DreamerV3, the same RL backbone as XIPER:
\begin{enumerate}
    \item VIPER~\cite{escontrela2024video}: a same-domain video prediction reward model.
    \item XAIL: a same-domain adversarial imitation learning baseline~\cite{peng2021amp,torabi2018generative} with DreamerV3, extended to the cross-domain setting using our translation model to map expert demonstrations into the agent domain.
    \item TPIL-Dv3: a cross-domain adversarial imitation learning method with a domain confusion loss~\cite{stadie2017third,choi2024domain} for domain-invariant discrimination, also with DreamerV3.
\end{enumerate}
All simulation agents learn from raw pixels without access to ground-truth state information or task rewards. All results are averaged over three independent runs.

\subsection{Performance of Cross-domain Video Model}
\label{sec:performance_video}
We begin by evaluating the cross-domain video prediction model, which directly determines the quality of the reward signal. More accurate predictions yield more informative rewards and better downstream agent performance.

\subsubsection{Domain Translation}
Since XIPER's reward signal depends on translating agent observations into the expert domain, translation quality is critical. We evaluate the translation model on expert trajectories, which are out-of-distribution relative to the random data it was trained on.

\begin{figure}[!htb]
    \centering
    \includegraphics[width=0.75\linewidth]{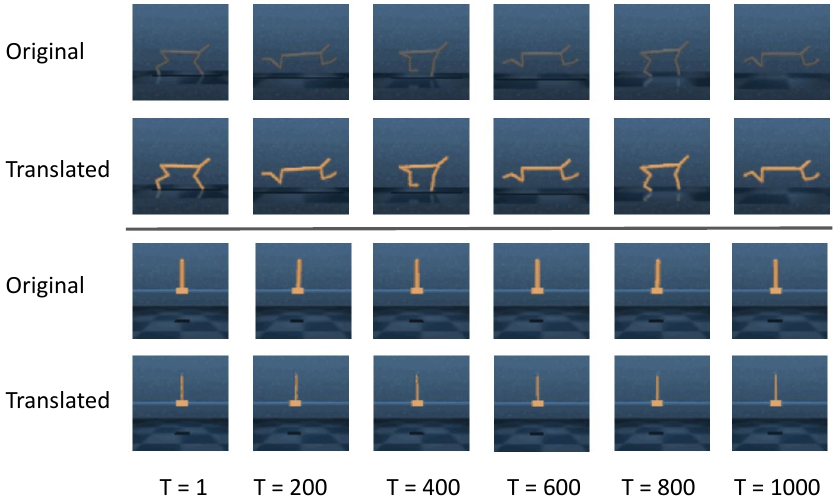}
    \caption{Expert trajectories in the agent domain translated into the expert domain. Although trained on randomly collected data, the model generalizes well to expert trajectories (OOD).}
    \label{fig:expert_translate}
\end{figure}

Despite being trained only on randomly collected observations, the model generalizes well to expert trajectories: in~\cref{fig:expert_translate}, gray agents in Cheetah-run are accurately translated to orange, and thick poles in Cartpole-balance are correctly resized.

To quantify this generalization, we report FID~\cite{heusel2017gans} scores of the translation model across various data subsets of Cheetah-run in~\cref{fig:fid_reward_models}~(a). FID measures the distance between two image distributions, where lower scores indicate greater similarity. Recall that the translation model is trained only on randomly collected observations. We evaluate three settings by measuring the FID between translated outputs and a target distribution: (R2R) translating random gray observations (agent's domain) and comparing to random orange (expert's domain), serving as the in-distribution baseline; (E2E) translating expert gray observations and comparing to expert orange, measuring OOD generalization; and (E2R) translating expert gray observations and comparing to random orange, serving as a reference to highlight the distribution gap between expert and random trajectories. R2R achieves a very low FID, confirming strong in-distribution performance. E2E remains relatively low, suggesting the model generalizes well to expert inputs despite being trained only on random data, likely because random observations provide broad coverage of the observation space. E2R is substantially higher than E2E, confirming that the gap is not due to poor translation quality but rather the inherent distributional difference between expert and random trajectories.

\subsubsection{Cross-domain Video Prediction Rollout}
\begin{figure}[!htb]
    \centering
    \includegraphics[width=0.65\linewidth]{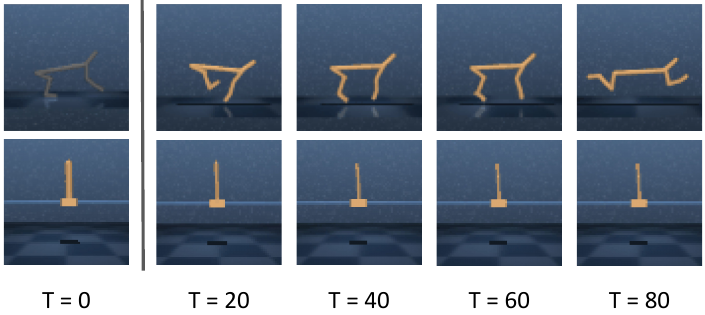}
    \caption{Predicted expert rollout of the cross-domain video prediction model, given an initial agent observation.}
    \label{fig:rollout}
\end{figure}

To evaluate the full cross-domain prediction pipeline, we initialize the model with an agent observation and generate a rollout of future expert observations. As shown in~\cref{fig:rollout}, the model successfully predicts expert-like trajectories from agent-domain inputs: it generates running motions of the orange expert given a gray agent's initial state, and produces balanced pole behaviors starting from the thick-pole variant. These predictions accurately capture expert dynamics, demonstrating the effectiveness of the combination of translation and prediction components.

\subsubsection{Using Likelihoods as Rewards}
\begin{figure}[!htb]
    \centering
    \includegraphics[width=0.75\linewidth]{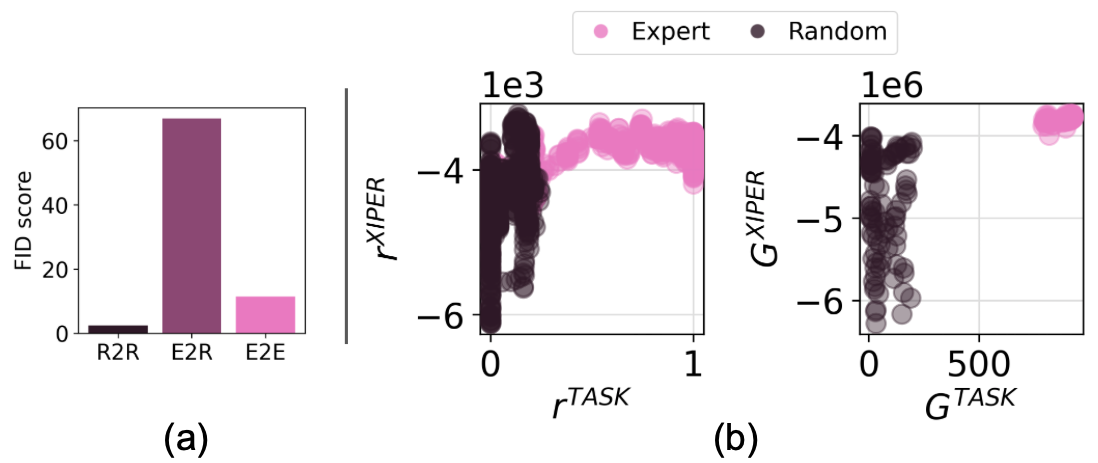}
    \caption{(a): FID scores on different data subsets of Cheetah-run, evaluated with the pretrained domain translation model. Lower FID indicates better translation quality. (b): Step-wise rewards (left) and episode returns (right) comparing XIPER and ground-truth task rewards on 100 random and expert trajectories.}
    \label{fig:fid_reward_models}
\end{figure}

To assess whether XIPER's likelihood-based rewards ($r^{\text{XIPER}}$) reflect true task performance, we compare them against ground-truth rewards ($r^{\text{TASK}}$) on 100 trajectories from random and expert policies in the Cheetah-run task~(\cref{fig:fid_reward_models}~(b)). XIPER assigns higher rewards to expert trajectories and lower rewards to random ones, with step-wise rewards and episode returns ($G^{\text{XIPER}}$ vs $G^{\text{TASK}}$) both correlating closely with ground-truth task rewards. This confirms that XIPER's prediction likelihoods serve as meaningful reward signals.

\subsection{XIPER Performance}
We now evaluate whether optimizing XIPER's learned rewards leads to meaningful task performance. Agents are trained solely with XIPER's likelihood-based reward signals and evaluated using ground-truth task rewards after 10M environment steps. An exception is TPIL-Dv3, which is trained for only 5M steps as it converges to a local optimum early in training.

\begin{figure*}[!htb]
    \centering
    \includegraphics[width=1\linewidth]{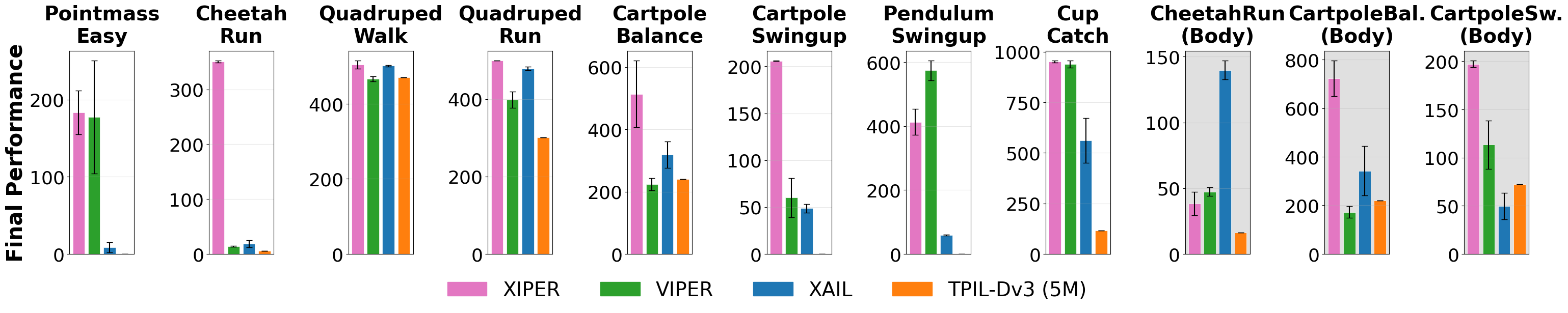}
    \caption{Final performance across all DMC tasks. XIPER outperforms all baselines in 9/12 tasks.}
    \label{fig:all_tasks}
\end{figure*}

In~\cref{fig:all_tasks}, we see across both suites, XIPER consistently achieves higher returns than baselines in most tasks. This result highlights the effectiveness of XIPER's reward model in the absence of ground-truth task feedback.

Notably, VIPER performs reasonably well in the DMC Color Suite despite lacking domain adaptation, suggesting that video prediction models can sometimes generalize over simple appearance shifts such as color changes. However, its performance collapses in the DMC Body Suite, where morphological changes create a larger domain gap. Both adversarial baselines, XAIL and TPIL-Dv3, perform poorly across both suites with unstable learning curves, confirming the known instability and data inefficiency of adversarial imitation learning even when paired with a strong RL backbone.

\subsection{Ablations}
\label{sec:ablation}

We examine two key design choices on three DMC Color tasks: the exploration coefficient $\beta$ and translation model quality.

\subsubsection{Exploration Coefficient $\beta$}
The performance of agents trained with varying exploration bonuses ($\beta$) from~\cref{eq:reward} is shown in~\cref{fig:ablations} (left three). While eliminating the exploration bonus ($\beta=0$) improves performance on certain tasks, it substantially degrades results on others, such as Cheetah-run. Overall, $\beta=1$ yields the most consistent performance across all tested tasks.

\subsubsection{Translation Model Quality}
\begin{figure}[!htb]
    \centering
    \includegraphics[width=0.8\linewidth]{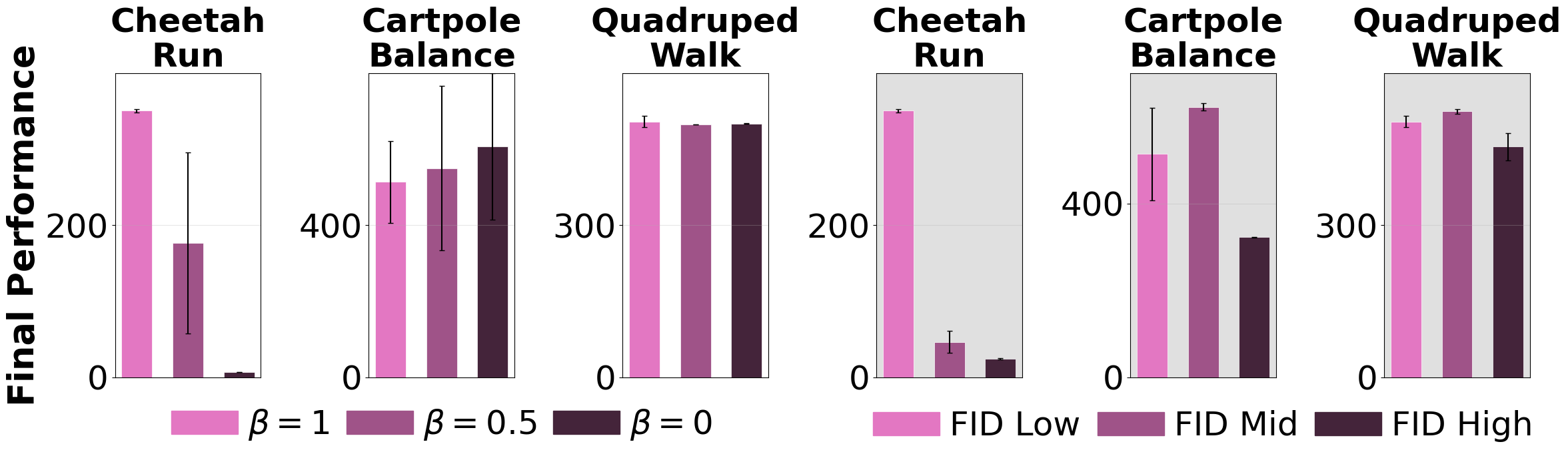}
    \caption{XIPER performance under different exploration coefficients $\beta$ and translation model qualities (FID) on DMC Color tasks. Lower FID and $\beta=1$ yield robust overall performance.}
    \label{fig:ablations}
\end{figure}

Since translation quality directly determines the fidelity of the reward signal, we train translation models of varying quality (measured by FID scores) and evaluate the resulting agents (right three in~\cref{fig:ablations}). High FID consistently yields poor performance across all tasks. While mid-range FID can match or exceed low FID on forgiving locomotion tasks, it catastrophically fails on precision-sensitive tasks, where even a modest FID degradation causes complete collapse on Cartpole Balance. This asymmetric sensitivity suggests that translation errors compound critically when precise state estimation is required. Low FID therefore represents the safest and most reliable choice.

\subsection{Sim2Real Analysis}
\label{sec:sim2real}
A compelling use case for XIPER is sim-to-real transfer, where expert demonstrations are only available in simulation. To analyze this potential, we examine XIPER on the LynxReach dataset using the Lynx Robot\footnote{\url{https://www.lynxmotion.com}}, where a robotic arm must reach a designated target from a random initial position. Rather than training a full RL agent, we analyze whether XIPER can generate meaningful reward signals for real-robot observations given only simulated expert videos. XIPER's reward model is trained in two stages: first, a domain translation model is trained on randomly collected simulation and real-robot data; second, a video prediction model is trained only on expert videos collected in simulation.

\subsubsection{Visual Translation Performance}
\begin{figure}[!htb]
    \centering
    \includegraphics[width=0.9\linewidth]{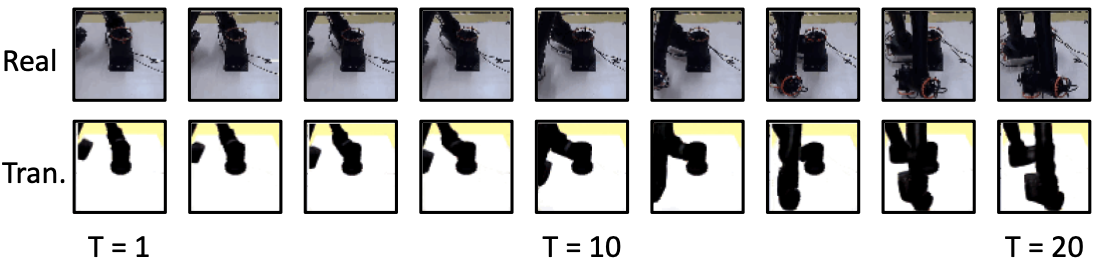}
    \caption{Sim2Real translation examples. Real robot expert trajectories (top) are translated into the simulation domain (bottom).}
    \label{fig:s2r_translation}
\end{figure}

We first evaluate the translation model's ability to map real-world observations into the simulation domain. As shown in~\cref{fig:s2r_translation}, the model successfully filters out task-irrelevant visual information such as cabling and hardware components, producing clean simulation-style images. Notably, despite being trained on randomly collected data, the model generalizes effectively to expert trajectories.

\subsubsection{Reward Signal Validation}
\begin{figure}[!htb]
    \centering
    \includegraphics[width=0.8\linewidth]{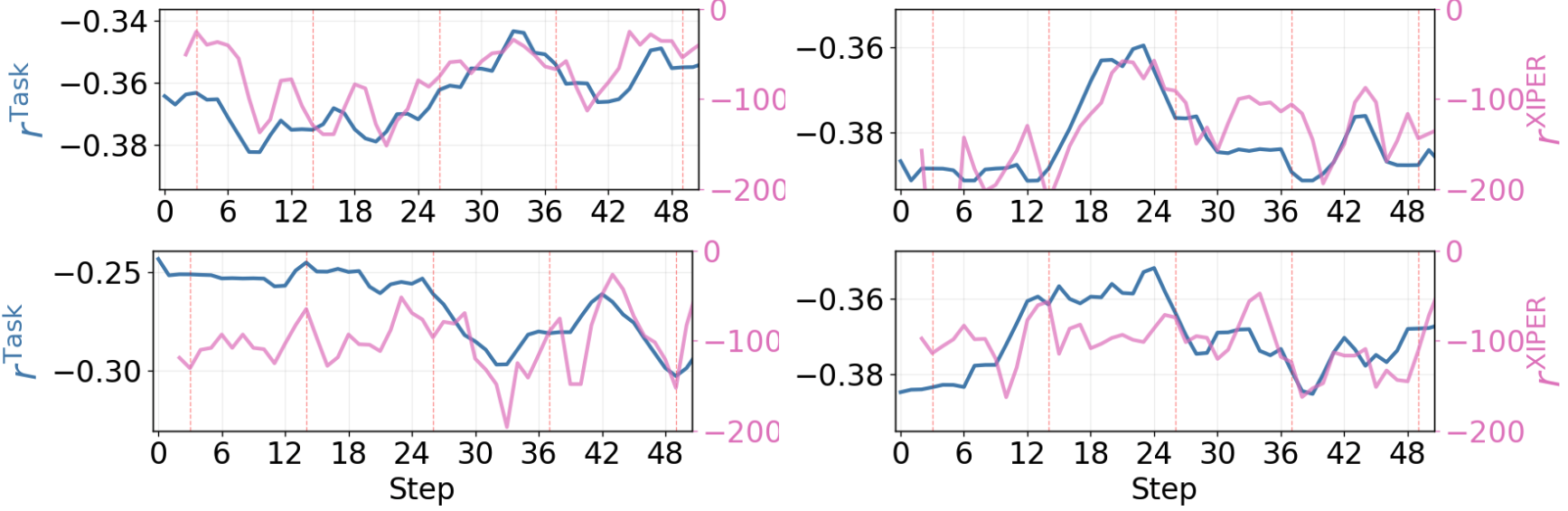}
    \caption{XIPER rewards vs.\ ground-truth task rewards on random real-robot episodes. The two signals exhibit consistently similar trends.}
    \label{fig:real_xiper_rewards}
\end{figure}

Having established translation quality, we next assess whether the resulting reward signals are semantically meaningful. Specifically, we apply the pretrained XIPER reward model (using Line 6 in~\cref{alg:viper}) to real-robot episodes and compare its outputs ($r^{\text{XIPER}}$) against ground-truth task rewards ($r^{\text{TASK}}$). 
As shown in~\cref{fig:real_xiper_rewards}, XIPER rewards correlate well with ground-truth rewards, demonstrating that simulated expert videos can provide a meaningful supervisory signal for real-robot observations. While this analysis does not involve training a full RL agent, it suggests that XIPER has the potential to bridge the sim-to-real gap for reward generation, motivating future work on end-to-end real-robot training.

\section{Conclusion}
\label{sec:conclusion}
We introduced XIPER, a cross-domain reward model that enables RL agents to learn from expert videos without access to environment rewards or ground-truth state information. XIPER maps agent observations into the expert domain and uses video prediction likelihood as a reward signal, guiding agents to reproduce expert behaviors across visual and morphological domain gaps. Experiments on the DMC Color and Body suites demonstrate that XIPER consistently outperforms adversarial and same-domain baselines, and our sim-to-real analysis shows that XIPER's reward signal correlates well with ground-truth task rewards on real-robot observations, suggesting its potential to bridge the sim-to-real gap without manual annotation.

Despite these promising results, two limitations remain. First, training the translation model solely on randomly collected data can be a bottleneck for tasks requiring precise, long-horizon behaviors, as random exploration may not provide sufficient coverage. Future work could address this through advanced exploration strategies~\cite{ma2024explorllm,park2025flow} or joint optimization of the translation model and RL policy. Second, while we validated XIPER's reward quality on a real-robot dataset, closing the loop with online RL training on a physical platform remains an important next step. We hope XIPER motivates future research into leveraging cross-domain videos as a scalable source of supervision for RL agents.

\section*{Acknowledgements}
We would like to thank Prof. Mark Hoogendoorn for his helpful guidance during this project. We also thank SURF (www.surf.nl) for the support in using the National Supercomputer Snellius. This work used the Dutch national e-infrastructure with the support of the SURF Cooperative using grant no. EINF-81853.

%
%
%
\bibliographystyle{splncs04}
\bibliography{ref}
%

\end{document}